\documentclass{article}

\usepackage{spconf,graphicx}
\graphicspath{ {./images/} }
\usepackage{subcaption}

\title{Which Prosodic Features Matter Most for Pragmatics?}
\name{Nigel G. Ward, Divette Marco, Olac Fuentes}
\address{University of Texas at El Paso \\ Computer Science Department \\ El Paso, Texas, 79912, USA}

\begin{document}
\ninept
\maketitle

\begin{abstract}
We investigate which prosodic features matter most in conveying prosodic functions.
We use the problem of predicting human perceptions of pragmatic similarity among utterance pairs
to evaluate the utility of prosodic features of different types. 
We find, for example, that duration-related features
are more important than pitch-related features, and that utterance-initial features are
more important than utterance-final features. 
Further, failure analysis indicates that modeling using pitch features only 
often fails to handle important pragmatic functions, and suggests  
that several generally-neglected acoustic and prosodic features 
are pragmatically significant, including nasality and vibrato.  
These findings can guide future basic research in prosody, and
suggest how to improve speech synthesis evaluation, among  
other applications.
\end{abstract}





\section{Motivation}

We ask: What prosodic features matter most in the expression of pragmatic functions? 

We choose to focus on pragmatic functions because of their importance in emerging
scenarios for speech technology, such as dialog systems involving interpersonal sensitivity or 
deployed in situated robots \cite{marge-espy-wilson-ward-csl}. 
Prosody is well-known to have important roles in 
conveying many pragmatic functions 
\cite{barth2009prosody,lameris2023,li2024spontaneous}. 

Our question may seem dated.  
Certainly it is now irrelevant
 for any short-term project 
for which there is adequate training data.
In such cases, rather than agonize over which features to use,
we can just use everything available, 
leaving it to the machine learning algorithm to exploit them effectively.
This strategy is especially well-suited to the use of pretrained models, 
which can encode much prosodic information  \cite{guan-ting-slim}. 

Nevertheless, attempting to answer this question could  serve: 

\noindent{\bf a)} To support system development, when lacking sufficient training data
to build a model from scratch.  Developers can use some knowlege of
which prosodic features generally matter most, to help them
build a starting-point architecture or model, 
which can then be fine-tuned on available data.

\noindent{\bf b)} To support prioritization, for psycholinguists and others doing basic research,
of research directions and questions, and for applied researchers, for example for
choosing what to focus on to describe some understudied language,
the nature of some communication disorder,
or some aspect of sociolinguistic variation. 

\noindent{\bf c)} To support the design of control parameters to
serve as ``knobs'' for human-in-the-loop specification or post-editing
of the prosody of speech synthesizer output \cite{iliescu2024controllable}.

\noindent{\bf d)}~ To enable better evaluation of the quality of the prosody output of a generative system,
for practical needs, including: 

\noindent{\bf d1)}~ The evaluation of speech synthesizer output \cite{wagner-beskow-slim}. 
A quick survey of the papers on speech synthesis at Interspeech 2023 
shows that, while most discuss prosody, the features considered were mostly pitch and duration, with only 10\% mentioning intensity and only one voicing properties. 
Even when specifically targeting better expressivity, conversational style, and better prosody
\cite{huang2023-slim,deng2024concss-slim,heffernan2024}, 
designers of metrics, 
lacking true knowledge of what matters, 
may fall back on what's familiar, namely pitch and sometimes duration. 

\noindent{\bf d2)} Building better speech-to-speech translation systems, 
where there is increasing interest in faithfully conveying more than just the lexical content \cite{seamless2023-slim,translatotron3}.  
Knowledge of which prosodic features matter
most can support the design of better loss functions.

\noindent{\bf d3)} Evaluating the power of discrete and other learned representations
\cite{leyuan-qu}.

\noindent {\bf d4)} The design of better speech codecs.  
Compression algorithms have been traditionally designed to maximize intelligibility and naturalness, 
but for use in interactive communication, 
preservation of certain prosodic features is also likely important. 

\noindent{\bf d5)} Designing interpretable feedback.  This can be for people wishing to learn how to communicate more effectively in business or in relationships, or for special populations
such as second language learners, 
those in rehabilitation after a stroke, and children with speech or language pathologies.

Over the decades, there have been numerous investigations of which 
prosodic features matter most for various specific purposes,
mostly in classification and regression tasks.  These include 
emotion recognition \cite{batliner10,gemaps-slim,vlasenko2024comparing}, 
 classification of linguistic structures such as tones, boundaries, and accents
\cite{ryant14,batliner99,batliner01},
 estimating judgments of language learners' accentedness, intelligibility, and other properties \cite{coutinho16-slim},
 prediction of turn-taking actions \cite{skantze-tt-review},
 language modeling \cite{timelm},
 speaker identification \cite{ferrer10},
 language identification \cite{ng09}, and
 detecting clinical conditions \cite{weed2020acoustic}.
Many interesting things have been found, and this body of work has  
greatly influenced the features included in prosodic-feature toolkits 
 \cite{gemaps-slim,midlevel-toolkit23}. 
However, none of this work has addressed pragmatics other than incidentally.

In this paper we use ``pragmatics" in a broad sense,
to include all aspects 
of interaction in dialog that go beyond the lexical semantic meaning.  
These are diverse: in dialog, people frequently show enthusiasm, make clarifications, 
cue action, clarify, criticize, praise, introduce new topics, 
yield the turn to the other, and so on.

\section{Methods}

Our interest is the prosody of pragmatic functions in general,
not for any one specific task. 
We accordingly chose to study prosody in the context of a general problem: estimating the perceived
pragmatic similarity between pairs of utterances.  
(This can be seen as a generalization of attempts to 
model how humans perceive similarity for intonation
\cite{hermes98,reichel-similarity,nocaudie16}, and of modeling
similar expressivity \cite{heffernan2024}).
Our working assumption is that a set of prosodic features that can support such estimates, across a wide variety of data,
is likely to be useful for many applications involving pragmatic functions.  
Thus, we leverage this model-building problem to glean insight into what aspects of prosody matter. 

\subsection{Data}

We exploit a dataset recently collected for another purpose \cite{ward-marco}.
This  consists of pairs of utterances, each with an
assessed pragmatic similarity value, on a continuous scale from 1 to 5,
based on the average rating of 6 to 9 human judges. 
There are 458 pairs in American English, our main focus for this paper, and 235 Northern Mexican Spanish pairs. 

Each utterance pair consists of a seed utterance extracted from a recorded dialog and a 
subsequent re-enactment of that utterance. 
Re-enactments were done under various conditions designed to create a variety of degrees of similarity. 
Critically, the dialogs were recorded with diverse scenarios to broaden the coverage of pragmatic functions
\cite{dral2023}, and within these, the seeds were also selected for diversity \cite{ward-marco}.
Thus the coverage is likely far broader than seen in any  single-genre corpus. 
Examples of the diversity of pragmatic functions appear below. 
 

In preliminary analysis we noted an interesting 
property of this data set: 
some feature distributions differ between the re-enactments and the seeds.
In particular the re-enactments tend to have less variation in the pitch features,
and to be louder and more creaky.

\subsection{Features}

We wanted a set of features that was broad in coverage, 
robust for dialog data, generally perceptually relevant, and simple,
to enable easy interpretation of the results. 

Specifically, we chose Avila's \cite{avila-ward23} adaptation of selected 
Midlevel Toolkit \cite{midlevel-toolkit23} features to tile utterances.
This set included 10 base features: 
 intensity, lengthening, creakiness, speaking rate, peak disalignment 
 (mostly late peak), 
cepstral peak prominence smoothed (CPPS), an
inverse proxy for breathy voice, and four pitch features, namely
measures of perceived pitch highness, pitch lowness, pitch wideness, and pitch narrowness.
While this feature set is far from ideal, it is suitable for this exploration.
Uniquely, it was designed to capture the prosody of pragmatic functions ---
unlike prosodic feature sets designed for paralinguistic properties, music,
or general signal processing --- 
and it was designed to be robust to microprosody and various phenomena of conversation.
At the same time, it is flawed.  Like other feature sets, none of its component features
is simultaneously fully robust, fully corresponding to perception, and fully accurate. 
For example, as the speaking rate feature is based on spectral flux, 
its values can be affected by diverse ancillary and confounding factors, 
including the presence of creaky voice, and the CPPS feature correlates only roughly 
with perceptions of breathiness.  
Nevertheless, when used for statistical and modeling purposes over sufficient data,
the features in this set can be useful, 
as seen by their utility in numerous basic and applied studies \cite{me-cup}.

Each base feature is normalized per track to be roughly speaker-independent.  
We then use average values over each of ten non-overlapping windows,
that span fixed percentages of its duration: 0–5\%, 5–10\%, 10–20\%, 20–30\%, 30–50\%, and symmetrically out to 100\%. 
This representation is not suited to syllable- or word-bound 
prosodic phenomena, but can roughly represent the sorts of overall levels and
contours that are  often associated with pragmatic functions.

This set is simplistic, and in particular includes nothing relating to time-sequence modeling,
notably no temporal deltas or functionals, but our working assumption is that we can still learn from it. 


\subsection{Models and Prediction Results}

While our main aim in this paper is to analyze feature importance,
modeling pragmatic similarity is a problem of importance in its own right 
\cite{ward-marco,me-interspeech24},
for example, for use cases d1--d3 above, so this subsection focuses on that perspective. 

Our primary metrics for model quality are correlations between the systems' similarity
estimates and the human judgments.  We also computed MSE, and the results were consistent. 
Our primary train/test split was between judgments collected in Sessions 1 and 2, a month apart. 
We also did experiments using 10-fold cross-validation across all the data, and the results were
similar. 

The models used are as follows: Euclidean Distance is a re-implementation of \cite{avila-ward23}.  
In this all features are weighted equally (after each is z-normalized).  
 For the next three models, the inputs were the 100 feature deltas,
that is, the feature values for the seed minus the values for the reenactment. 
(Performance using instead the absolute differences was always somewhat lower,
as one might expect for models blind to the seed-reenactment distinction.)
For the KNN Regression Model, $k$ was 50. 
For the Random Forest Regression there were 100 trees. 
The "selected HuBert cosine" uses the cosine similarity between feature representations
consisting of 103 Hubert layer-24 features selected to maximize performance on the training data \cite{me-interspeech24}.  
We note that none of these models is very sophisticated ---
lacking dynamic time warping or other alignment methods, use of average- or max-pooling, 
non-linear or configurational compositions of features, and so on --- 
but are adequate for exploratory purposes. 


\begin{table}[t]
\begin{center}
    \begin{tabular}{lrrr}
    & Correlation \\
    \hline\\[-1.5ex]
    Euclidean Distance & --0.33 ~~~~ \\
    Linear Regression   & 0.44 ~~~~ \\
    KNN Regression    & 0.58 ~~~~ \\
    Random Forest Regression   & 0.70  ~~~~ \\ 
    cosine over selected HuBert & 0.74 ~~~~
    \end{tabular}
\end{center}
\caption{Pearson's correlation between each models' predictions and the human judgments.}
\label{tab:model-EN2-correlations}
\end{table}


The results are seen in Table \ref{tab:model-EN2-correlations}.  
First, we see the usual trade-off between model simplicity/explainability and performance. 
More interestingly, we see that the best designed-features model, with random forest regression,
is doing almost as well as the pretrained features model.
This indicates that the penalty for using designed features is small,
and use of a more sophisticated decision model might close the gap.
Further, examining the correlations separately for 
pairs which were lexically different
and for pairs which were lexically-identical,
performance on the former was near-random, but 0.80 for the latter, as good as with 
pretrained features \cite{me-interspeech24}. 

\section{Feature-Importance Analyses}

Given a set of features and a task, there are many ways to
measure the importance of  feature sets and subsets \cite{batliner99}.  
We accordingly investigated using three methods. First, we 
simply computed the Pearson's correlation between each of the 100 features 
and the target judgments.  Second we examined how much each feature contributed to
the performance of the Random Forest Regression Models; specifically, 
we obtained the feature importance values in terms of impurity decrease
for each fold and averaged these. 
Third, we did subset and ablation studies, examining performance when 
including only, or when excluding only, various feature types.

The implications noted below are all multiply supported,
so, to save space, we present only a selection of the evidence.  
However we note that there is no consistent ranking of features, 
as seen in Table \ref{tab:importance-per-feature-type}.
This is not surprising; rather, the existence of feature types with low correlations but relatively high 
importance indicates that the features are not independent,
and instead, as often noted \cite{me-cup}, specific configurations of features
likely bear specific meanings.
We also note that the features are highly redundant: ablating any specific
type only slightly reduces the performance. 

\subsection{Most Important Feature Types}

\begin{figure}[th]
\centering
\includegraphics[clip=true, trim=4mm 0mm 10mm 13mm, width=\columnwidth]{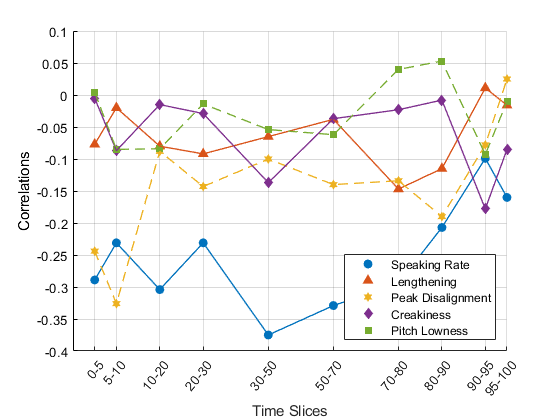}
\caption{Single-feature correlations between the judgments 
and the deltas for five of the most informative feature types, 
across both Session 1 and Session 2 data. 
The X-axis represents the regions, defined by fixed percentages of the utterance duration.   }
\label{fig:EN-feature-correlation}
\end{figure}

Table \ref{tab:importance-per-feature-type} shows results per feature type,
and Figure \ref{fig:EN-feature-correlation} displays the correlations 
for five informative feature types,
We draw three implications:
1) Duration features are important, with speaking rate the top feature by every measures.
Interestingly, while lengthening is strongly anti-correlated with speaking rate, we it still has some independent value, increasing the performance over speaking rate alone by 0.02 (correlation with human judgments). 
2) The value of the pitch features is low.
This was not entirely a surprise,  
because the reasons that pitch is popular  --- 
being salient, easy to visualize, familiar from music, 
relatively easy to measure, and historically important --- 
do not imply actual utility, and 
we suspected that the self-evident importance of pitch features for modeling read speech 
may not carry over to dialog.  
However the importance of the pitch features was surprisingly small.  As this is the first study to actually measure the value of 
prosodic features for pragmatic functions, the last word is yet to be written, 
but we can conclude at least that pitch features do not deserve the exclusive respect 
that they often get.
3) The least informative features overall are intensity and pitch narrowness.  

\begin{table}[t]
\begin{center}
    \begin{tabular}{lrr}
    \multicolumn{1}{c}{Feature} & \multicolumn{1}{c}{Importance} & \multicolumn{1}{c}{Correlation}\\
    \hline & \\[-1.8ex]
    speaking rate   & 43.7\% ~~& 0.64~~\\
    lengthening     & 20.9\% ~~& 0.54~~\\
    peak disalignment & 7.8\% ~~& 0.32~~\\
    CPPS            & 5.9\% ~~& 0.04~~\\
    pitch highness  & 4.1\% ~~& 0.12~~\\
    pitch narrowness & 4.0\% ~~& --0.06~~\\
    pitch wideness  & 3.9\% ~~& 0.00~~\\
    creakiness      & 3.5\% ~~& 0.14~~\\
    pitch lowness   & 3.3\% ~~& 0.13~~\\
    intensity       & 3.1\% ~~& --0.03~~\\
    4 pitch features\rule{0ex}{2.7ex}& 15.2\% ~~& 0.16~~\\
    all 10 features  & 100.0\%  ~~& 0.70~~\\ 
    \end{tabular}
\end{center}
\caption{Feature types, ordered by importance for the random forest regression model
and also showing performance of a model using features of this type alone.}
\label{tab:importance-per-feature-type}
\end{table}

\begin{figure}[b!]
\begin{center}
\includegraphics[clip=true, trim=7mm 0.5mm 1mm 7mm, width=\columnwidth]{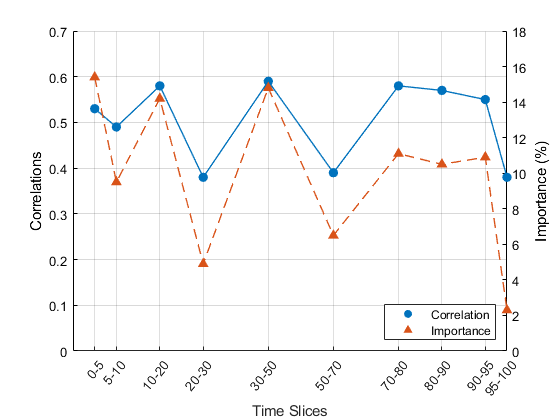}
\end{center}
\caption{Feature importance as a function of position (time slice):  on the left axis, performance of a model using only features at that position; on the right axis, summed importance in the random forest regression model.}
\label{fig:per-time}
\end{figure}
\subsection{Most Important Feature Positions}

Figure \ref{fig:per-time} summarizes some of the evidence regarding 
which feature positions matter most. 
It seems that the prosody about 70--90\% into the utterances 
is relatively informative.
Since this was especially true for peak 
disalignment, pitch wideness, pitch highness, and lengthening,
we suspect that this may relate to the common occurrence of various types of pitch peak 
(such as nuclear accents) at this point in many utterances. 
There were other interactions between feature type and position ---  
speaking rate is 
especially informative in the beginnings and middles of utterances, 
peak disalignment at the beginning, 
and the lengthening feature mostly toward the end ---  
but these did not seem to fall into any pattern. 
There is some evidence that the prosody early in utterances is 
relatively more important and the final prosody is the least important, 
but the latter may be overstated in the figure, 
as the 95-100\% features are not robust 
to variation in where the labelers marked utterance ends, 
and in these dialogs the utterances often trailed off.
Overall, however, we did not find any strong tendencies 
regarding which feature positions are most important. 


\section{Qualitative Analyses}

Our first qualitative analysis was a brief 
exploration of why pitch-only model performed so much worse than the all-feature model.
We examined a small sampling of pairs for which the predictions of the 
former were far more accurate than those of the latter. 
Several pragmatic functions were common in these pairs, mostly commonly 
positive/negative assessment, turn hold/yield,
and correction of a misunderstanding. 
The most common prosodic-acoustic properties present in these pairs, 
which also seemed to be involved in conveying these meanings,
included nasality and speaking rate variation.  
Saliently, each of these poorly-handled pairs had a synthesized-speech re-enactment.
From this we infer that the synthesizer used, 
namely Amazon Polly, is not able to effectively control (or even much vary, it seems) 
many of the prosodic characteristics that are important to human perception.  
As far as we know, this is also likely true for all synthesizers,
and we speculate that this is due in large part to the pitch-prioritizing loss functions
that they are trained to. 

Our second qualitative analysis explored the limitations of Avila's 100-feature set.
Although designed to be widely inclusive, it
did not quite support state-of-the-art performance, at least with the models tried. 
Again we did failure analysis, this time more thoroughly, 
examining 30 pairs for which the predictions of our best model,
the Random Forest Regression model, tested in 10-fold cross validation, 
diverged most from the human judgments. 

First we examined pairs which the model rated much higher than did the judges, 
looking for differences that our ears could hear
but that the model likely had missed. 
Almost all of these involved differences in nasality. 
Also common were differences in pause frequency, length, and location.
Other factors we noticed include, in rough order of frequency, 
words said with or without laughing,
the exact phonetic form of non-lexical utterances, such as {\it oh},
phonetic reduction including devoicing,
stressing of specific words, 
vibrato, falsetto,  non-lexical sighs, 
uses of glottal stops, ejectives, and strong harmonicity.
Speaking rate variations and 
breathiness were also common factors, 
even though the speaking rate and CPPS features were 
intended to cover for such perceptions.

Second, we examined pairs that the model rated much lower than the judges.  
One frequent factor was differences in pacing or pause placement that 
were not significant to our ears, but seemed to trip up the model. 
This suggests, unsurprisingly, that a model could do better with  some kind of alignment 
or max-pooling. 
Another factor was apparent individual- or, often, gender-based variant prosodic 
forms for conveying the same meaning. 
For example, in {\it oh my god, it's working} (female) and {\it yo, it's working} (male), 
where, in addition to the lexical differences, the former used vibrato, breathy and falsetto voice, 
and the male creaky voice, both conveyed excitement and matched well in nuance. 
Thus, while most aspects of spoken English are amenable to gender-agnostic modeling, 
this suggests that this strategy will not work well for pragmatics-related prosody. 


Third, we revisited these pairs and a sampling of pairs that were handled well, hoping to discover which pragmatic-function distinctions remain
problematic, even with the full feature set.  However, there were no
clear patterns: all the functions identified as problematic
for the pitch-only model were often handled well by the full model,
and of course, the magnitude of the divergences was much less 
than for the pitch-only model. 



Audio illustrating these points is available at https://www.cs.utep. edu/nigel/pros-prag/.

\section{Spanish}

Wondering which of the findings above might apply beyond English, 
we repeated most of the analyses using the Spanish data from \cite{ward-marco}. In brief, we found:
1) These features serve to predict pragmatic similarity fairly well for Spanish too 
(0.73 correlation, with 10-fold cross-validation).
2) The feature types with the highest correlations only partly overlapped those 
for English, with the top three being speaking rate, creakiness, and pitch wideness.  
3) Modeling using pitch alone was again far inferior to using all features,
but the penalty was less than for English (correlation of 0.41, with 10-fold cross-validation),
4) Utterance-final features were again the least informative, 
5) Models trained on one language and tested on the other performed reasonably well 
(e.g. Spanish trained on English: correlation 0.68), but not as well as language-specific models. 
6) Common features
lacking from the model but important for human perception of differences
again included nasality and devoicing.
7)  Humans often overlooked differences in creakiness, nasality, and breathiness, 
and these often seemed to reflect gender-specific patterns of use.



\section{Summary and Limitations}

We have reported the results of the first systematic study of
which prosodic features matter most for pragmatics.
While we can, of course provide no definitive answer to this question ---
the best feature set will always
depend on the task, language, speaker population and so on ---
this exploration has contributed:   

\begin{itemize}
\item a new method for evaluating the pragmatic adequacy of prosodic feature sets
\item some explainable models for predicting human judgments of pragmatic similarity 
\item a new method for discovering important but lacking features
\item indications for which features should be included in evaluation metrics for applications,
notably not only pitch features but also duration-related and voicing features
\item identification of pragmatic functions that are poorly handled by pitch-only feature sets, 
including making corrections, marking positive or negative feeling, and indicating turn hold/yield intentions
\item identification of features that are understudied but important for pragmatics, and thereby deserving of 
further study, notably nasality and vibrato
\end{itemize}

While we have broken new ground, we note that this study has numerous limitations, 
including the small data sizes, 
the simplicity of the features and modeling, 
the lack of coverage of all genres of dialog, and
the focus on American English. 
Further work is needed. 

\subsection*{Acknowledgment} 

This work was supported in part by National Science Foundation award 2348085 and the AI Research Institutes program of the NSF and by the Institute of Education Sciences, U.S. Department of Education through Award \# 2229873 -- National AI Institute for Exceptional Education. 

\parskip 0pt

\bibliographystyle{IEEEbib}
      \bibliography{bib}

\begin{thebibliography}{10}

\bibitem{marge-espy-wilson-ward-csl}
Matthew Marge, Carol Espy-Wilson, Nigel~G. Ward, et~al.,
\newblock ``Spoken language interaction with robots: Research issues and
  recommendations,''
\newblock {\em Computer Speech and Language}, vol. 71, 2022.

\bibitem{barth2009prosody}
Dagmar Barth-Weingarten, Nicole Deh{\'e}, and Anne Wichmann,
\newblock {\em Where Prosody Meets Pragmatics},
\newblock Brill, 2009.

\bibitem{lameris2023}
Harm Lameris, Joakim Gustafsson, and {\'E}va Sz{\'e}kely,
\newblock ``Beyond style: Synthesizing speech with pragmatic functions,''
\newblock in {\em Interspeech}, 2023, pp. 3382--3386.

\bibitem{li2024spontaneous}
Weiqin Li, Peiji Yang, et~al.,
\newblock ``Spontaneous style text-to-speech synthesis with controllable
  spontaneous behaviors based on language models,''
\newblock in {\em Interspeech}, 2024.

\bibitem{guan-ting-slim}
Guan-Ting Lin, Chi-Luen Feng, et~al.,
\newblock ``On the utility of self-supervised models for prosody-related
  tasks,''
\newblock in {\em IEEE Workshop on Spoken Language Technology (SLT)}, 2022, pp.
  1104--1111.

\bibitem{iliescu2024controllable}
Dan~Andrei Iliescu, Devang S~Ram Mohan, Tian~Huey Teh, and Zack Hodari,
\newblock ``Controllable prosody generation with partial inputs,''
\newblock in {\em IEEE ICASSP}, 2024, pp. 11916--11920.

\bibitem{wagner-beskow-slim}
Petra Wagner, Jonas Beskow, et~al.,
\newblock ``Speech synthesis evaluation: State-of-the-art assessment and
  suggestion for a novel research program,''
\newblock in {\em Proceedings of the 10th Speech Synthesis Workshop (SSW10)},
  2019.

\bibitem{huang2023-slim}
Wen-Chin Huang, Benjamin Peloquin, et~al.,
\newblock ``A {{Holistic Cascade System}}, {B}enchmark, and {{Human Evaluation
  Protocol}} for {{Expressive Speech-to-Speech Translation}},''
\newblock in {\em {ICASSP}}, 2023.

\bibitem{deng2024concss-slim}
Yayue Deng, Jinlong Xue, et~al.,
\newblock ``Con{CSS}: Contrastive-based context comprehension for
  dialogue-appropriate prosody in conversational speech synthesis,''
\newblock in {\em IEEE ICASSP}, 2024, pp. 10706--10710.

\bibitem{heffernan2024}
Kevin Heffernan, Artyom Kozhevnikov, et~al.,
\newblock ``Aligning speech segments beyond pure semantics,''
\newblock in {\em Findings of the Association for Computational Linguistics},
  2024, pp. 3626--3635.

\bibitem{seamless2023-slim}
Lo{\"\i}c Barrault, Yu-An Chung, et~al.,
\newblock ``Seamless: Multilingual expressive and streaming speech
  translation,''
\newblock {\em arXiv preprint arXiv:2312.05187}, 2023.

\bibitem{translatotron3}
Eliya Nachmani, Alon Levkovitch, Yifan Ding, Chulayuth Asawaroengchai, Heiga
  Zen, and Michelle~Tadmor Ramanovich,
\newblock ``Translatotron 3: Speech to speech translation with monolingual
  data,''
\newblock in {\em IEEE ICASSP}, 2024, pp. 10686--10690.

\bibitem{leyuan-qu}
Leyuan Qu, Taihao Li, et~al.,
\newblock ``Disentangling prosody representations with unsupervised speech
  reconstruction,''
\newblock {\em IEEE/ACM Transactions on Audio, Speech, and Language
  Processing}, 2023.

\bibitem{batliner10}
Anton Batliner, Stefan Steidl, Bjorn Schuller, et~al.,
\newblock ``Whodunnit: Searching for the most important feature types
  signalling emotion-related user states in speech,''
\newblock {\em Computer Speech and Language}, vol. 25, pp. 4--28, 2011.

\bibitem{gemaps-slim}
Florian Eyben, Klaus~R. Scherer, et~al.,
\newblock ``The {G}eneva minimalistic acoustic parameter set ({GeMAPS}) for
  voice research and affective computing,''
\newblock {\em IEEE Transactions on Affective Computing}, vol. 7, pp. 190--202,
  2016.

\bibitem{vlasenko2024comparing}
Bogdan Vlasenko, Sargam Vyas, et~al.,
\newblock ``Comparing data-driven and handcrafted features for dimensional
  emotion recognition,''
\newblock in {\em IEEE ICASSP}, 2024, pp. 11841--11845.

\bibitem{ryant14}
Neville Ryant, Malcolm Slaney, Mark Liberman, Elizabeth Shriberg, and Jiahong
  Yuan,
\newblock ``Highly accurate {M}andarin tone classification in the absence of
  pitch information,''
\newblock in {\em Proceedings of Speech Prosody}, 2014.

\bibitem{batliner99}
Anton Batliner, Jan Buckow, Richard Huber, Volker Warnke, Elmar N{\"o}th, and
  Heinrich Niemann,
\newblock ``Prosodic feature evaluation: {B}rute force or well designed,''
\newblock in {\em Proc. 14th Int. Congress of Phonetic Sciences}, 1999, vol.~3,
  pp. 2315--2318.

\bibitem{batliner01}
Anton Batliner, Jan Buckow, Richard Huber, Volker Warnke, Elmar N\"oth, and
  Heinrich Niemann,
\newblock ``Boiling down prosody for the classification of boundaries and
  accents in {G}erman and {E}nglish,''
\newblock in {\em Eurospeech}, 2001, pp. 2781--2784.

\bibitem{coutinho16-slim}
Eduardo Coutinho, Florian H{\"o}nig, et~al.,
\newblock ``Assessing the prosody of non-native speakers of {E}nglish: Measures
  and feature sets,''
\newblock in {\em Conference on Language Resources and Evaluation (LREC 2016)},
  2016, pp. 1328--1332.

\bibitem{skantze-tt-review}
Gabriel Skantze,
\newblock ``Turn-taking in conversational systems and human-robot interaction:
  A review,''
\newblock {\em Computer Speech \& Language}, vol. 67, pp. 101178, 2021.

\bibitem{timelm}
Nigel~G. Ward, Alejandro Vega, and Timo Baumann,
\newblock ``Prosodic and temporal features for language modeling for dialog,''
\newblock {\em Speech Communication}, vol. 54, pp. 161--174, 2011.

\bibitem{ferrer10}
Luciana Ferrer, Nicolas Scheffer, and Elizabeth Shriberg,
\newblock ``A comparison of approaches for modeling prosodic features in
  speaker recognition,''
\newblock in {\em IEEE ICASSP}, 2010, pp. 4414--4417.

\bibitem{ng09}
Raymond W.~M. Ng, Tan Lee, Cheung-Chi Leung, Bin Ma, and Haizhou Li,
\newblock ``Analysis and selection of prosodic features for language
  identification,''
\newblock in {\em IEEE International Conf. on Asian Language Processing}, 2009,
  pp. 123--128.

\bibitem{weed2020acoustic}
Ethan Weed and Riccardo Fusaroli,
\newblock ``Acoustic measures of prosody in right-hemisphere damage: A
  systematic review and meta-analysis,''
\newblock {\em Journal of Speech, Language, and Hearing Research}, vol. 63, no.
  6, pp. 1762--1775, 2020.

\bibitem{midlevel-toolkit23}
Nigel~G. Ward,
\newblock ``Midlevel prosodic features toolkit (2016-2023),''
\newblock {h}ttps://github.com/nigelgward/midlevel, 2023.

\bibitem{hermes98}
Dik~J. Hermes,
\newblock ``Auditory and visual similarity of pitch contours,''
\newblock {\em Journal of Speech, Language, and Hearing Research}, vol. 41, pp.
  63--72, 1998.

\bibitem{reichel-similarity}
Uwe~D. Reichel, Felicitas Kleber, and Raphael Winkelmann,
\newblock ``Modelling similarity perception of intonation,''
\newblock in {\em Interspeech}, 2009, pp. 1711--1714.

\bibitem{nocaudie16}
Olivier Nocaudie and Corine Ast{\'e}sano,
\newblock ``Evaluating prosodic similarity as a means towards {L2} teacher’s
  prosodic control training,''
\newblock {\em Speech Prosody 2016}, pp. 26--30, 2016.

\bibitem{ward-marco}
Nigel~G. Ward and Divette Marco,
\newblock ``A collection of pragmatic-similarity judgments over spoken dialog
  utterances,''
\newblock in {\em Linguistic Resources and Evaluation Conference}, 2024.

\bibitem{dral2023}
Nigel~G. Ward, Jonathan~E. Avila, Emilia Rivas, and Divette Marco,
\newblock ``Dialogs re-enacted across languages, version 2,''
\newblock Tech. {R}ep. UTEP-CS-23-27, {U}niversity of Texas at El Paso,
  Department of Computer Science, 2023.

\bibitem{avila-ward23}
Jonathan~E. Avila and Nigel~G. Ward,
\newblock ``Towards cross-language prosody transfer for dialog,''
\newblock in {\em Interspeech}, 2023.

\bibitem{me-cup}
Nigel~G. Ward,
\newblock {\em Prosodic Patterns in English Conversation},
\newblock Cambridge University Press, 2019.

\bibitem{me-interspeech24}
Nigel~G. Ward, Andres Segura, Alejandro Ceballos, and Divette Marco,
\newblock ``Towards a general-purpose model of perceived pragmatic
  similarity,''
\newblock in {\em Interspeech}, 2024.

\end{thebibliography}
     
\end{document}